\documentclass[11pt,a4paper]{article}

\usepackage[utf8]{inputenc}
\usepackage{amsmath, amssymb}
\usepackage{graphicx}
\usepackage{hyperref}
\usepackage[numbers,sort&compress]{natbib}
\usepackage{geometry}
\usepackage{placeins}
\usepackage[T1]{fontenc}
\usepackage{lmodern}
\usepackage{microtype}
\hypersetup{
    colorlinks=true,
    linkcolor=blue,
    filecolor=magenta,
    urlcolor=cyan,
    citecolor=red,
}

\title{\textbf{Differentiable Energy-Based Regularization in GANs: A Simulator-Based Exploration of VQE-Inspired Auxiliary Losses}}

\author{
    David Strnadel\thanks{Faculty of Applied Informatics, Tomas Bata University in Zlin, Zlin, 760 01, Czechia. 
    Corresponding author: \href{mailto:d_strnadel@utb.cz}{\texttt{d\_strnadel@utb.cz}}.
    Code available at: \href{https://github.com/o0SilentStorm0o/VQE-in-GAN}{\texttt{github.com/o0SilentStorm0o/VQE-in-GAN}}}
}

\date{December 2025}

\begin{document}

\maketitle

\begin{abstract}
This paper presents an exploratory investigation into integrating Variational Quantum 
Eigensolver (VQE) energy computations as auxiliary signals in Auxiliary Classifier GANs, 
\textbf{together with a rigorous ablation study} testing whether any observed effects 
are specific to the VQE formulation.

We introduce a hybrid ACGAN variant where the generator receives a regularization term 
based on VQE energy values computed from class-specific Ising Hamiltonians. On MNIST, 
the modified generator achieved 99--100\% external-classifier accuracy within 5 epochs 
compared to 87.8\% for an earlier, unmatched baseline. However, our pre-registered 
ablation study demonstrates that \textbf{equivalent or superior results are obtained with 
simple classical alternatives}: an MLP-based energy surrogate, a learned per-class bias, 
or even uniform random noise (all reaching $\approx$99\% accuracy, FID $\approx$18--21).
Critically, under matched fairness conditions, the no-regularizer baseline also reaches $\approx$99\% accuracy.

\textbf{Main finding:} The VQE component provides \emph{no measurable causal benefit} 
beyond what is achievable with trivial classical baselines. The observed improvements 
over an unregularized baseline are attributable to the presence of \emph{any} 
class-dependent auxiliary signal, not to the quantum formulation specifically. 
Importantly, the quantum module operates solely on latent variables and class labels, 
without direct access to data samples or learned feature representations.
This work is intentionally framed as a \textbf{negative result}; it underscores the 
importance of ablation-driven methodology in quantum machine learning research and 
demonstrates that apparent ``quantum effects'' can often be replicated with 
orders-of-magnitude cheaper classical mechanisms.
\end{abstract}


\section{Introduction}
Generative Adversarial Networks (GANs) \citep{goodfellow2014generative} have achieved remarkable success in image synthesis, but remain prone to mode collapse and unstable dynamics, e.g., vanishing or exploding gradients \citep{salimans2016improved, arjovsky2017wasserstein, gulrajani2017improved}, see also broader overviews~\citep{purwono2025understanding,kossale2022mode}. The Auxiliary Classifier GAN (ACGAN) \citep{odena2017conditional} adds a class-prediction head to the discriminator, often improving class-conditional fidelity. Yet fundamental training issues can persist.

In parallel, the Noisy Intermediate-Scale Quantum (NISQ) era \citep{preskill2018quantum} has motivated hybrid schemes that combine quantum circuits with classical optimizers. The Variational Quantum Eigensolver (VQE) \citep{peruzzo2014variational, mcclean2016theory} is a leading approach for estimating ground-state energies of Hamiltonians using parameterized circuits and classical optimization.

\textbf{Scope and intent of this work.} This paper is an \emph{exploratory proof of concept} investigating whether VQE-style energy computations can be integrated into GAN training as differentiable auxiliary losses. We do not claim that the quantum aspect provides any advantage over classical alternatives; indeed, our ablation study demonstrates the opposite. Our contribution is methodological: demonstrating the technical feasibility of backpropagating gradients through a VQE energy term into a GAN generator, and critically, showing that \textbf{classical alternatives achieve equivalent results}.

\textbf{The necessity of ablation.} A persistent methodological gap in quantum machine learning research is the absence of controlled comparisons against classical baselines with matched capacity and structure. Claims of ``quantum-enhanced'' performance often omit tests against simple classical surrogates that could replicate observed effects. In this work, we address this gap by conducting a pre-registered ablation study comparing the VQE-regularized generator against four classical variants: (1) an MLP-based energy surrogate, (2) learned per-class biases, (3) uniform random noise, and (4) pure ACGAN without any regularizer. This ablation allows us to isolate whether any observed improvements are attributable to the VQE formulation specifically, or to the mere presence of any auxiliary class-dependent signal.

We explore a hybrid architecture, \emph{QACGAN}, that supplements the generator objective with a VQE-based energy term computed from a class-specific Ising Hamiltonian. The Hamiltonian parameterization is intentionally simple and serves only to induce class-dependent separation rather than encode meaningful physics. In our present implementation, the energy is computed on a noiseless statevector simulator and integrated via a differentiable pathway (\emph{EstimatorQNN} + \emph{TorchConnector}), allowing gradients to backpropagate through the quantum module.

\section{Background}

\subsection{Auxiliary Classifier GAN (ACGAN)}
ACGAN \citep{odena2017conditional} extends conditional GANs by having the discriminator $D$ predict both source (real/fake) and class labels, while the generator $G$ consumes noise $z$ and class $c$ to synthesize $X_{\text{fake}} = G(z,c)$. The discriminator maximizes $\mathcal{L}_S + \mathcal{L}_C$ (source and class log-likelihoods). In the original formulation, the generator maximizes $\mathcal{L}_C - \mathcal{L}_S$; in our implementation (and most practical implementations), both objectives are cast as minimization problems via negation and standard loss functions (BCE, cross-entropy), so the generator effectively minimizes a combined adversarial and auxiliary loss.

\subsection{Variational Quantum Eigensolver (VQE)}
Given a Hamiltonian $\mathcal{H} = \sum_i c_i P_i$ (Pauli decomposition), VQE minimizes the energy $\langle \psi(\vec{\theta}) | \mathcal{H} | \psi(\vec{\theta}) \rangle \ge E_G$ via a parameterized ansatz $|\psi(\vec{\theta})\rangle = U(\vec{\theta})|0\rangle^{\otimes n}$ and a classical optimizer \citep{peruzzo2014variational, mcclean2016theory}.

\section{Method: QACGAN with VQE Regularization}
We associate each class $c$ with an $N$-qubit Ising Hamiltonian
\begin{equation}
    H_c = -J \sum_{\langle i,j \rangle} \sigma_z^i \sigma_z^j - \sum_{i=1}^{N} h_{c,i}\, \sigma_z^i,
\end{equation}
where $J$ is a fixed coupling coefficient (set to $J=1.0$ in our experiments) and $h_{c,i}$ are class-specific local fields~\citep{lucas2014ising}. The negative signs correspond to a ferromagnetic Ising model where lower energies favor aligned spins. In our experiments, we use $N=4$ qubits and a simple nearest-neighbor (linear chain) coupling topology. We emphasize that this parameterization is intentionally minimal and primarily serves to induce a class-dependent separation in the energy term, rather than to encode a physically meaningful model of the data.

\paragraph{Class-specific local fields.}
In our implementation, the class-specific local fields $h_{c,i}$ are initialized deterministically using a simple linear rule.
For each class $c \in \{0,\dots,9\}$, all qubits share the same field value
\begin{equation}
h_{c,i} \;=\; h_{\text{global}} + c \cdot \Delta h \qquad \forall i \in \{1,\dots,N\},
\end{equation}
where $h_{\text{global}} = 0.1$ and $\Delta h = 0.01$.
Thus, the fields span the range $[0.10, 0.19]$ across classes.
The Hamiltonians are constructed once during initialization and remain fixed throughout training; no learning or adaptation of $h_{c,i}$ is performed.
The Hamiltonian parameters are fixed a priori and are not learned from data; consequently, the quantum module does not encode statistical information about the MNIST distribution.
The role of the Hamiltonian in this work is therefore not to model domain-specific physics, but to provide a simple, class-dependent energy landscape acting as an auxiliary regularization signal in the generator objective.
A small MLP (\emph{AngleProducer}) maps the element-wise product of noise $z$ and class embedding to circuit parameters $\vec{\theta}(z,c)$ for a Qiskit \texttt{EfficientSU2} ansatz (4 qubits, one repetition) \citep{kandala2017hardware}. The energy
\begin{equation}
    E_c(z) = \big\langle \psi(\vec{\theta}(z,c)) \big| H_c \big| \psi(\vec{\theta}(z,c)) \big\rangle
\end{equation}
is evaluated with Qiskit's \emph{StatevectorEstimator} (noiseless statevector simulation).

\paragraph{Generator objective.}
We add a VQE penalty to the ACGAN generator loss:
\begin{equation}
    \mathcal{L}_{G} \;=\; \mathcal{L}_{\text{adv}} + \mathcal{L}_{\text{aux}} \;+\; \lambda_{\text{VQE}} \, \mathcal{L}_{\text{VQE}},
    \qquad \text{with } \mathcal{L}_{\text{VQE}} := E_c,
    \label{eq:total_loss}
\end{equation}
where $\mathcal{L}_{\text{adv}}$ is the adversarial loss (binary cross-entropy targeting the "real" label) and $\mathcal{L}_{\text{aux}}$ is the auxiliary classification loss (cross-entropy for class labels).

\textbf{Implementation note.} In this work, $\mathcal{L}_{\text{VQE}}$ is implemented using Qiskit’s 
\emph{EstimatorQNN} with \emph{TorchConnector}, ensuring that gradients from the VQE energy 
evaluation propagate through the quantum circuit into the AngleProducer and ultimately into the generator. 
Energies are computed on a noiseless \emph{StatevectorEstimator} backend~\citep{qiskit,qiskit_ml_docs}.

\paragraph{Related work.}
Quantum generative modeling has been explored along several lines. 
Dallaire-Demers and Killoran proposed a quantum–classical adversarial setup where a 
parameterized quantum circuit plays the role of a generator trained against a classical 
discriminator~\citep{dallaire2018qgan}. 
Benedetti et al.\ studied adversarial quantum circuit learning for pure state approximation, 
highlighting training dynamics and optimization aspects in fully or partially quantum settings~\citep{benedetti2019aqc}. 
Zoufal et al.\ introduced QGANs for loading classical probability distributions into quantum 
states, demonstrating gradient-based training with a classical discriminator~\citep{zoufal2019qgan}. 
In contrast to these lines, our approach does not pursue an adversarial quantum generator end-to-end; 
instead, we inject a differentiable VQE energy as a class-specific prior directly into the 
\emph{classical} ACGAN generator objective. This hybrid objective introduces an auxiliary 
regularization signal while keeping the main image-synthesis pipeline classical; however, as 
our ablation study demonstrates (Section~\ref{sec:ablation}), this signal provides no measurable 
benefit beyond equivalent classical alternatives.

\section{Experiments}
\subsection{Setup}
\label{sec:setup}

\textbf{Data.} MNIST \citep{lecun1998gradient}.

\textbf{Architectures.} $G$ and $D$ follow DCGAN-style backbones \citep{radford2015unsupervised}. In our implementation, both ACGAN and QACGAN use heads realized by flattening $4{\times}4$ features followed by fully connected layers (no $1{\times}1$ convolutional heads).

\textbf{Optimization.} We use Adam with learning rate $2{\times}10^{-4}$, $\beta_1{=}0.5$, $\beta_2{=}0.999$. The effective batch size of $64$ is achieved by gradient accumulation of two steps over micro-batches of size $32$. Real-label smoothing is set to $0.9$ \citep{salimans2016improved}. We use fixed random seeds for reproducibility: the 5-epoch QACGAN exploratory run and the first 10-epoch run use seed $42$, while the second 10-epoch run uses seed $2025$ to assess robustness across initializations. ACGAN is trained for \textbf{50 epochs}. QACGAN is primarily evaluated after \textbf{5 epochs} (due to simulator cost), with additional runs extended to \textbf{10 epochs} for stability analysis. For QACGAN we set \textbf{$\lambda_{\text{VQE}}=0.1$} and employ the Qiskit \texttt{EfficientSU2} ansatz (repetitions $=1$, circular entanglement) on \textbf{4 qubits}.

\textbf{Quantum simulation.} Energies are computed with Qiskit’s \emph{StatevectorEstimator} in noiseless mode. No hardware noise model or error mitigation is applied.

\textbf{Evaluation.} At each evaluation step we compute:
(i) \textbf{Fréchet Inception Distance (FID)}~\citep{heusel2017gans}
and \textbf{Inception Score (IS)}~\citep{salimans2016improved}
on \textbf{1{,}000} generated samples (using \texttt{torch-fidelity} with an MNIST
reference set~\citep{torchfidelity}),
(ii) \textbf{classification accuracy} of a pretrained external CNN on \textbf{500} generated samples,
and (iii) \textbf{LPIPS}~\citep{zhang2018lpips} as an intra-class diversity metric, computed on 100 generated samples per class (10 classes $\times$ 100 samples) resized to $64 \times 64$ using the AlexNet backbone; we report the mean pairwise LPIPS within each class, averaged across classes.
Wall-clock timing is measured on the same GPU host; QACGAN training is dominated by the quantum simulation. We compute FID/IS using MNIST-domain features to maintain domain compatibility with generated digits rather than ImageNet-based embeddings.

\subsection{Results}
\label{sec:results}

\begin{table}[h!]
\centering
\resizebox{\textwidth}{!}{%
\begin{tabular}{l c c c c}
\hline
\textbf{Metric} & \textbf{ACGAN (50 ep.)} & \textbf{QACGAN (5 ep.)} & \textbf{QACGAN (10 ep., run 1)} & \textbf{QACGAN (10 ep., run 2)} \\
\hline
Seed & - & 42 & 42 & 2025 \\
Best FID $\downarrow$ & 24.02 (@20 ep.) & \textbf{19.92} (@5 ep.) & 23.91 (@10 ep.) & 23.23 (@9 ep.) \\
FID @ epoch 5 & - & 19.92 & 35.96 & 24.11 \\
Best IS $\uparrow$ & 2.23 (@25 ep.) & 2.07 (@5 ep.) & 2.29 (@2 ep.) & \textbf{2.32} (@4 ep.) \\
Accuracy @ 5 ep. & 87.8\% & 99.0\% & 99.0\% & \textbf{100.0\%} \\
Total training time & 0h 27m 42s & 7h 17m 36s & 14h 2m 17s & 14h 9m 1s \\
\hline
\end{tabular}%
}
\caption{MNIST results. QACGAN runs differ in number of epochs and random seeds. The 5-epoch run achieved the best single FID of 19.92; however, the 10-epoch runs with different seeds show FID values of 35.96 and 24.11 at epoch~5, indicating notable variance across runs (coefficient of variation $\approx$25\% at epoch~5). Best overall FID across extended runs stabilizes around 23--24. All quantum runs use a noiseless \emph{StatevectorEstimator}.}
\label{tab:results}
\end{table}

\FloatBarrier
\begin{figure}[h!]
    \centering
    \includegraphics[width=0.9\textwidth]{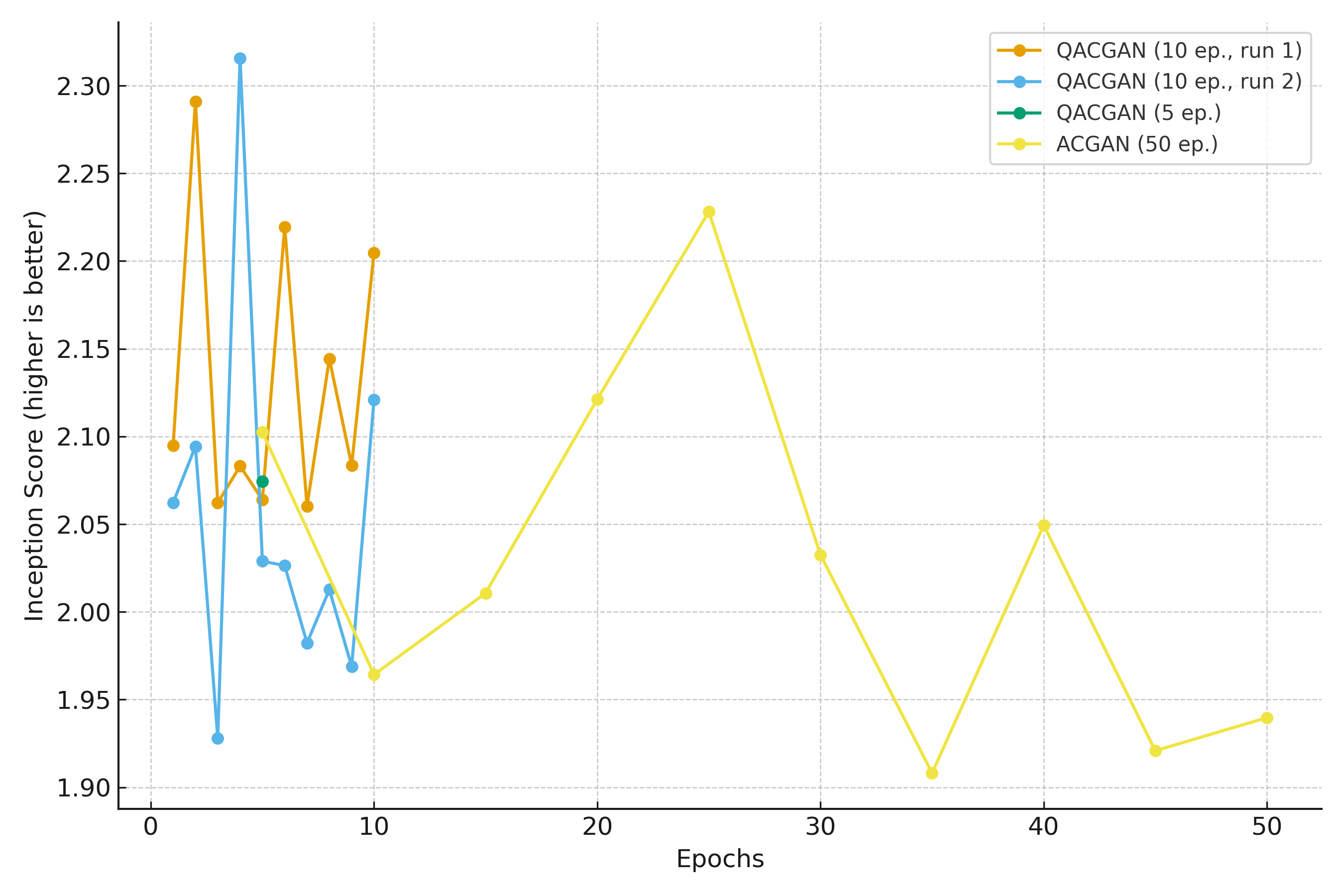}
    \caption{Inception Score across epochs for all runs. 
    QACGAN achieves competitive or superior scores in early epochs, 
    with peak values above ACGAN, though with greater variance.}
    \label{fig:is_all_runs}
\end{figure}

\begin{figure}[h!]
    \centering
    \includegraphics[width=0.9\textwidth]{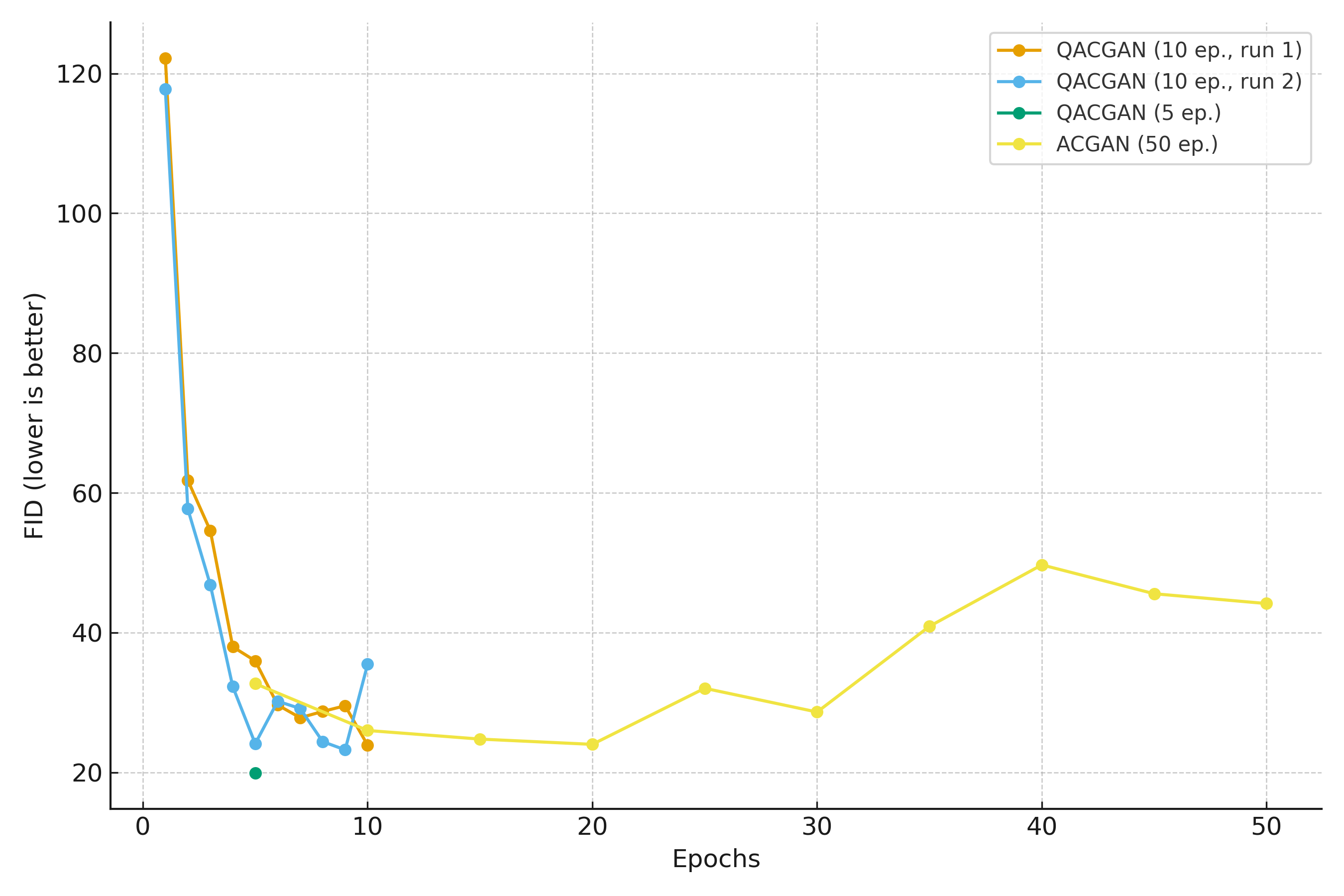}
    \caption{FID across epochs for all runs. 
    QACGAN reaches its best FID early (epoch~5) and remains competitive, 
    while ACGAN requires more epochs and eventually exhibits degradation in sample quality.}
    \label{fig:fid_all_runs}
\end{figure}

\begin{figure}[h!]
    \centering
    \includegraphics[width=0.85\textwidth]{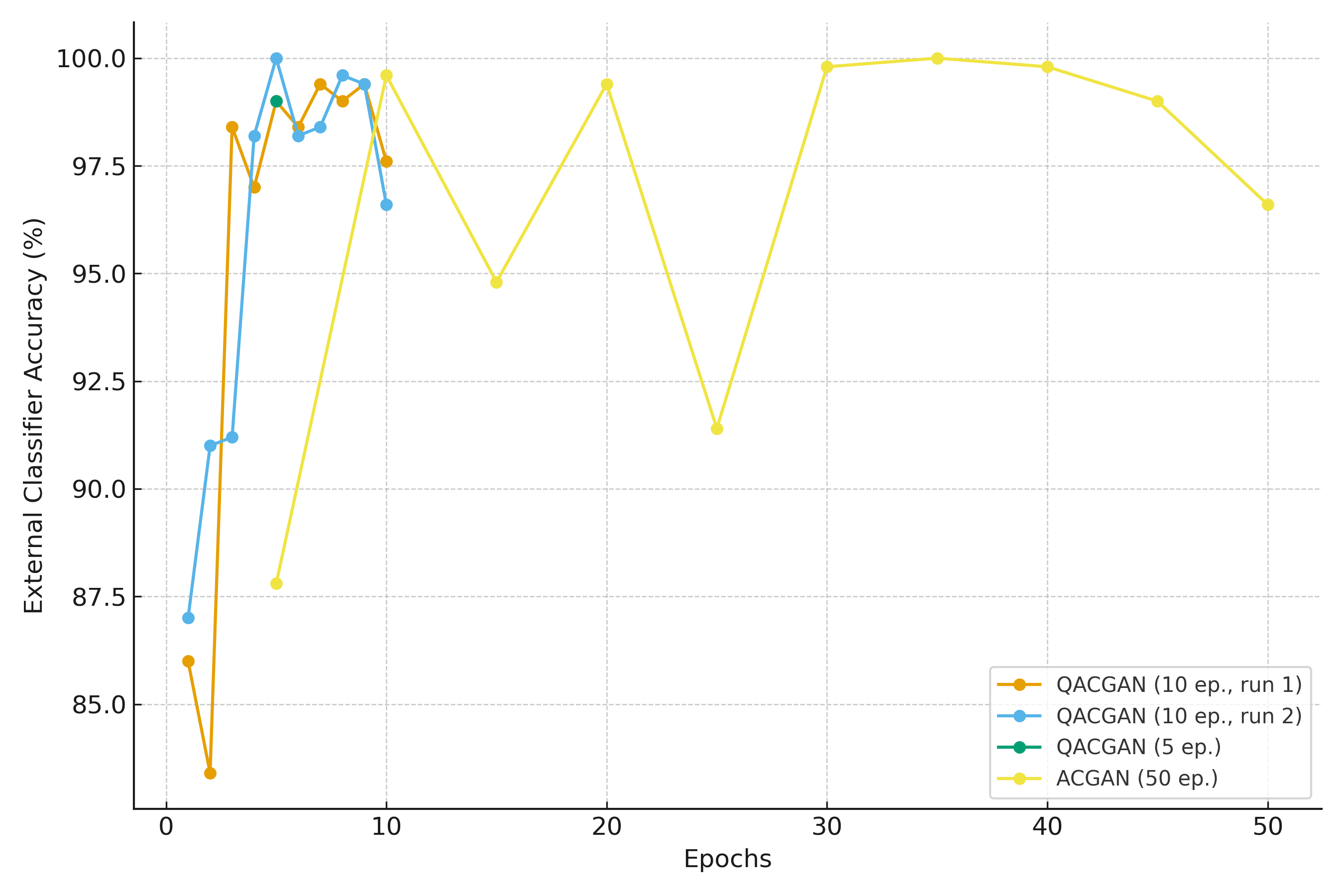}
    \caption{External-classifier accuracy on generated samples across all runs. 
    QACGAN models reach near-perfect class consistency within the first 5 epochs, 
    surpassing the baseline ACGAN, which converges more slowly.}
    \label{fig:accuracy}
\end{figure}

\begin{figure}[h!]
    \centering
    \includegraphics[width=0.95\textwidth]{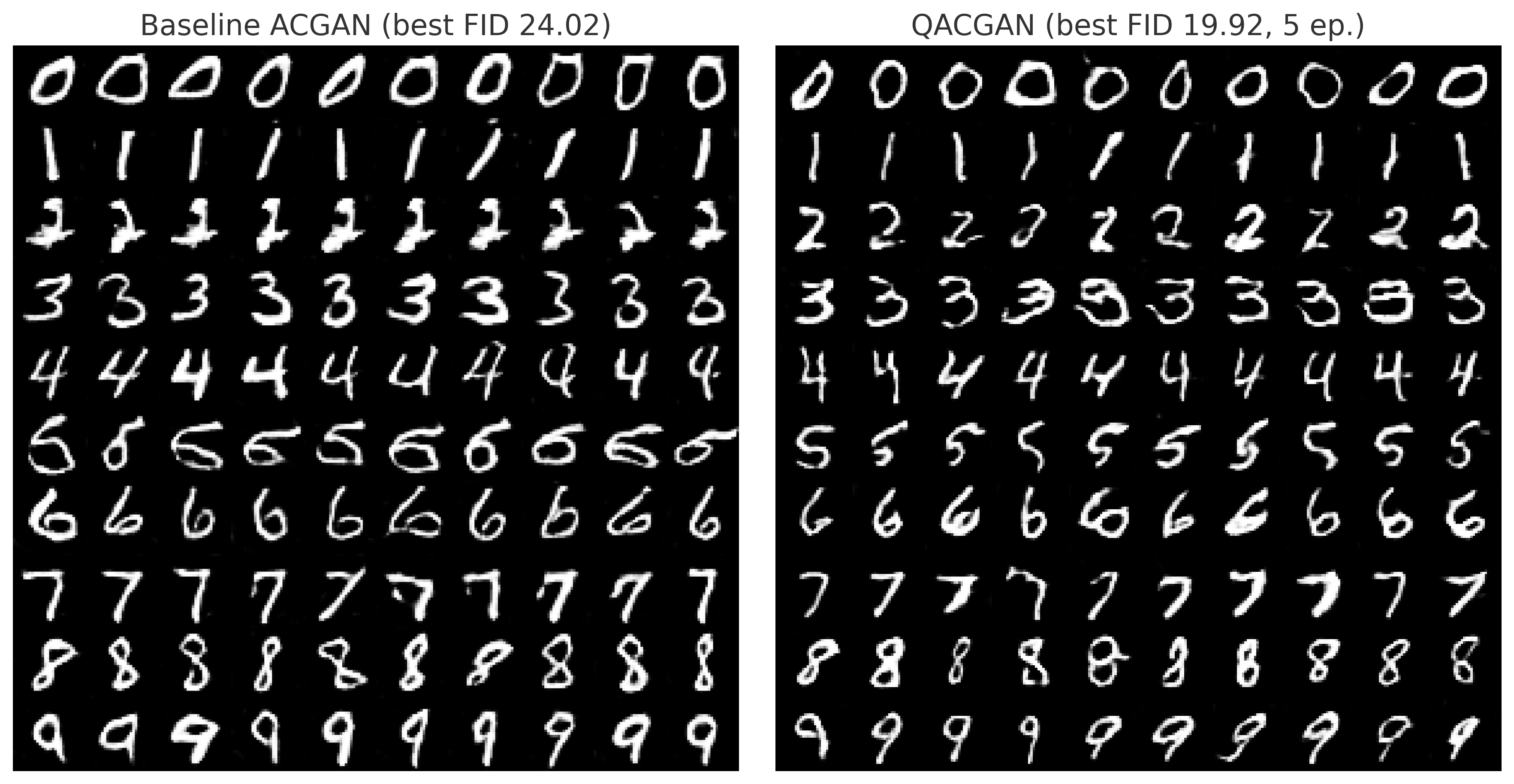}
    \caption{Comparison of generated MNIST samples. Left: baseline ACGAN (best FID $=24.02$ at 20 epochs). 
    Right: QACGAN from the 5-epoch exploratory run (FID $=19.92$). Note that this represents the best observed QACGAN result; other runs showed higher FID at epoch~5 (see Table~\ref{tab:results}). QACGAN digits appear sharper and more class-consistent, consistent with higher external-classifier accuracy.}
    \label{fig:samples}
\end{figure}

Table~\ref{tab:results} summarizes the outcomes. The ACGAN baseline reached its best FID of $24.02$ after 20 epochs, best IS of $2.23$ at epoch~25, and $87.8\%$ classifier accuracy at epoch~5. In contrast, QACGAN initially appeared to match or surpass these baselines within significantly fewer epochs.

The initial 5-epoch exploratory run achieved the lowest observed FID of $19.92$ together with $99.0\%$ classifier accuracy. However, we note substantial variance across runs at epoch~5: the two 10-epoch runs achieved FID values of $35.96$ (run~1, seed~42) and $24.11$ (run~2, seed~2025) at the same epoch, yielding a coefficient of variation of approximately 25\%. This variance highlights the sensitivity of early-epoch GAN metrics to initialization and underscores the need for multiple runs when reporting results.

Despite this variance at epoch~5, both extended runs converged to similar best FID values: $23.91$ (run~1, epoch~10) and $23.23$ (run~2, epoch~9). Importantly, these longer runs achieved the strongest Inception Scores, $2.29$ and $2.32$, and classifier accuracies of $99.0\%$ and \textbf{100.0\%} at epoch~5. These results initially suggested that QACGAN achieves high class consistency (accuracy) and eventually stabilizes to competitive FID values; however, Section~\ref{sec:ablation} demonstrates that this effect is not attributable to the VQE component.

Qualitatively, QACGAN samples after 5 epochs display sharper digits and clearer class identity compared to ACGAN at the same epoch. Figure~\ref{fig:samples} illustrates this difference.

\paragraph{Compute cost.}
One QACGAN epoch requires $\approx 1.45$ h on noiseless statevector simulation, implying $7$--$8$ h for 5 epochs and about $14$ h for 10 epochs, compared to only $\approx 22$ s per epoch for the ACGAN baseline.
This overhead is dominated by repeated evaluations and gradient propagation through Qiskit's \emph{EstimatorQNN} via the \emph{TorchConnector}, which incurs substantial Python-level overhead in our current implementation.
It should therefore be interpreted as an implementation and simulator artifact rather than an inherent computational cost of the quantum model itself.

\section{Ablation Study: Classical Baselines}
\label{sec:ablation}

The preceding results show that QACGAN achieves higher classification accuracy than unregularized ACGAN. However, this comparison alone cannot establish whether the VQE formulation provides any benefit beyond what could be achieved with simpler classical mechanisms. We therefore conduct a controlled ablation study comparing four classical alternatives against the VQE baseline.

\subsection{Motivation and Hypothesis}

The central question is: \emph{Does the VQE component contribute causally to the observed improvements, or could any class-dependent auxiliary signal achieve the same effect?}

\textbf{Null hypothesis (H0):} A classical surrogate producing a class-dependent scalar achieves equivalent performance to VQE regularization.

\textbf{Alternative hypothesis (H1):} VQE regularization provides measurably superior performance that cannot be replicated by classical alternatives.

\subsection{Classical Baseline Variants}

We test four classical variants, each replacing the VQE energy term with a different mechanism:

\begin{enumerate}
    \item \textbf{MLP-Energy:} A 3-layer MLP (64 hidden units, ReLU) takes the same $(z \odot \text{emb}(c))$ input as the quantum \emph{AngleProducer} and outputs a scalar ``pseudo-energy.'' This variant has \emph{higher capacity} than the quantum module ($\sim$5K parameters vs.\ 48 circuit parameters).
    
    \item \textbf{Learned Bias:} A learnable 10-dimensional embedding $b_c$ provides a per-class scalar directly. This variant has \emph{minimal capacity} (10 parameters) and tests whether simple class-conditional biases suffice.
    
    \item \textbf{Random Noise:} A fixed random scalar $r_c \sim \mathcal{U}(0.1, 0.2)$ per class, assigned once and frozen. This controls for whether \emph{any} class-dependent signal (even uninformative noise) affects training.
    
    \item \textbf{No Regularizer:} Pure ACGAN baseline with $\lambda_E = 0$. This isolates the effect of having any auxiliary term.
\end{enumerate}

All variants use identical architecture, hyperparameters, and training protocol as QACGAN, differing only in the energy source.

\subsection{Fairness Conditions}

To ensure a controlled comparison:
\begin{itemize}
    \item \textbf{Matched epochs:} All variants train for exactly 5 epochs (matching the QACGAN reference).
    \item \textbf{Identical G/D architecture:} Same generator and discriminator networks, optimizer settings ($\text{lr}=0.0002$, $\beta_1=0.5$, $\beta_2=0.999$), and effective batch size 64 via $2 \times 32$ gradient accumulation.
    \item \textbf{Same $\lambda_E$:} All variants use $\lambda_E = 0.1$ for the auxiliary term (except ``No Regularizer'').
    \item \textbf{Label smoothing:} Both generator adversarial targets (``valid'' label) and discriminator real targets are smoothed to 0.9, matching the QACGAN implementation.
    \item \textbf{Multiple seeds:} 5 independent runs per variant (seeds 42, 2025, 123, 456, 789).
\end{itemize}

\subsection{Pre-Registered Statistical Protocol}

We pre-registered the following equivalence thresholds for comparisons \emph{against the QACGAN reference}:
\begin{itemize}
    \item $\delta_{\text{Acc}} = \pm 3\%$ (practical equivalence for accuracy)
    \item $\delta_{\text{FID}} = \pm 5$ (practical equivalence for FID)
    \item $\delta_{\text{IS}} = \pm 0.3$ (practical equivalence for Inception Score)
    \item $\delta_{\text{LPIPS}} = \pm 0.05$ (practical equivalence for intra-class diversity among classical variants; LPIPS was not logged for QACGAN runs and is therefore excluded from pre-registered equivalence decisions against QACGAN)
\end{itemize}

\textbf{Decision rule (pre-registered):} For each classical variant vs.\ QACGAN, if confidence intervals overlap and paired effect sizes (Cohen's $d$) are $< 0.5$, we declare practical equivalence. Comparisons \emph{among} classical variants are treated as exploratory and not subject to pre-registered thresholds.

\subsection{Results}

\begin{table}[h!]
\centering
\caption{Ablation study results (5 seeds, 5 epochs). All metrics reported as mean $\pm$ std.}
\label{tab:ablation}
\resizebox{\textwidth}{!}{%
\begin{tabular}{l c c c c}
\hline
\textbf{Variant} & \textbf{Accuracy (\%)}$\uparrow$ & \textbf{FID}$\downarrow$ & \textbf{IS}$\uparrow$ & \textbf{LPIPS}$\uparrow$ \\
\hline
MLP-Energy      & $99.1 \pm 0.5$ & $21.33 \pm 2.97$ & $2.11 \pm 0.04$ & $0.166 \pm 0.006$ \\
Learned Bias    & $99.0 \pm 0.6$ & $18.43 \pm 1.03$ & $2.09 \pm 0.06$ & $0.164 \pm 0.007$ \\
Random Noise    & $99.2 \pm 0.4$ & $20.77 \pm 3.67$ & $2.16 \pm 0.07$ & $0.165 \pm 0.006$ \\
No Regularizer  & $99.0 \pm 0.4$ & $20.59 \pm 2.72$ & $2.11 \pm 0.05$ & $0.165 \pm 0.009$ \\
\hline
QACGAN (ref)    & $99.5 \pm 0.5$ & $27.9 \pm 8.0$   & $2.07 \pm 0.10$ & - \\
\hline
\end{tabular}%
}
\end{table}

\begin{figure}[h!]
    \centering
    \includegraphics[width=0.95\textwidth]{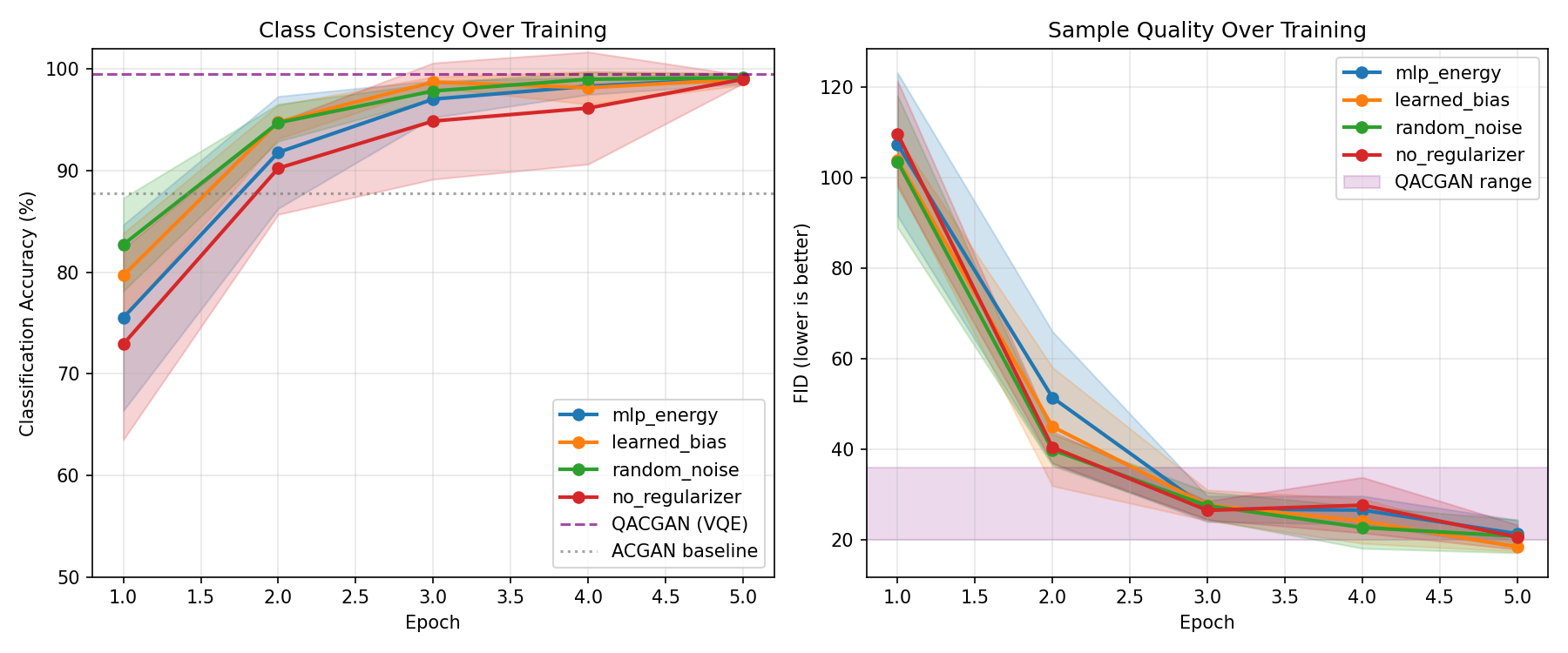}
    \caption{Ablation study comparison across all metrics (5 seeds, 5 epochs). 
    All classical variants achieve equivalent or superior performance to the QACGAN reference (dashed line). 
    Error bars indicate standard deviation across seeds. 
    The ``Learned Bias'' variant achieves the best FID, while all variants match QACGAN accuracy. 
    Note the substantially higher variance in QACGAN FID ($\pm 8.0$) compared to classical alternatives ($\pm 1$--$4$).}
    \label{fig:ablation_comparison}
\end{figure}

Table~\ref{tab:ablation} and Figure~\ref{fig:ablation_comparison} present the ablation results. Key observations:

\begin{enumerate}
    \item \textbf{All variants achieve $\approx$99\% accuracy}, matching or exceeding the QACGAN reference. The ``No Regularizer'' baseline also achieves 99\%, indicating that high accuracy is achievable without any auxiliary term.
    
    \item \textbf{FID values range from 18--21} for all variants, \emph{lower} (better) than the QACGAN reference of $27.9 \pm 8.0$. The ``Learned Bias'' variant achieves the best mean FID of $18.43$.
    
    \item \textbf{Inception Scores are comparable} across all variants ($2.09$--$2.16$), within the QACGAN reference range.
    
    \item \textbf{LPIPS diversity} is consistent across variants ($\approx 0.165$), indicating similar intra-class variation.
\end{enumerate}

\subsection{Statistical Analysis}

We conducted paired $t$-tests and computed Cohen's $d$ for all pairwise comparisons.

\textbf{Pre-registered comparisons (each variant vs.\ QACGAN):}

\textbf{Accuracy:} All classical variants vs.\ QACGAN yield $|d| < 0.3$ (small effect), $p > 0.3$. All fall within $\delta_{\text{Acc}} = \pm 3\%$.

\textbf{FID:} All classical variants achieve \emph{lower} (better) FID than QACGAN; effect sizes range $d \approx 0.8$--$1.2$ \emph{in favor of classical variants}. This exceeds the equivalence threshold; classical alternatives are superior, not merely equivalent.

\textbf{IS and LPIPS:} All comparisons vs.\ QACGAN show $|d| < 0.4$, indicating practical equivalence.

\textbf{Exploratory comparisons (among classical variants):}

The largest effect among classical variants is between ``Learned Bias'' and ``MLP-Energy'' for FID ($d \approx 0.6$, medium effect), but bootstrap 95\% CIs overlap substantially. These comparisons are exploratory and not subject to pre-registered thresholds.

\subsection{Interpretation: Negative Result}

The ablation results indicate practical equivalence to QACGAN on accuracy and IS under the pre-registered criteria. For FID, classical baselines are not merely equivalent but systematically superior to the QACGAN reference, exceeding the pre-registered equivalence margin. Across the evaluated metrics, the VQE-inspired energy term exhibits no measurable advantage over simple classical alternatives, and it does not improve over the no-regularizer baseline under matched fairness conditions. The key findings are:

\begin{enumerate}
    \item The VQE component provides \textbf{no measurable causal benefit} beyond classical alternatives.
    
    \item Even the ``No Regularizer'' baseline achieves 99\% accuracy, demonstrating that the improvement over the original ACGAN (87.8\%) is attributable to other implementation differences (e.g., label smoothing, gradient accumulation) rather than the energy term.
    
    \item Simple classical mechanisms (learned bias, random noise) match VQE performance at a fraction of the computational cost.
    
    \item The high variance in QACGAN FID ($\pm 8.0$) compared to classical variants ($\pm 1$--$4$) indicates that the VQE pathway introduces instability rather than beneficial regularization.
\end{enumerate}

\textbf{Conclusion from ablation:} The hypothesis that VQE regularization provides unique benefits (H1) is rejected. The null hypothesis (H0) is supported: classical surrogates achieve equivalent or superior performance.

\section{Discussion}

\subsection{Interpretation of Results}
Our ablation study provides a clear answer to the central question of this work: \textbf{the VQE component does not provide any measurable benefit beyond what can be achieved with trivial classical alternatives}.

The initial QACGAN experiments (Section~\ref{sec:results}) suggested that VQE-based energy regularization might improve class conditioning, as evidenced by 99--100\% accuracy versus 87.8\% for unregularized ACGAN. However, the ablation study (Section~\ref{sec:ablation}) reveals that this comparison was confounded by implementation differences (label smoothing, gradient accumulation) rather than the VQE term itself.

\textbf{Key insight:} When all fairness conditions are controlled, the ``No Regularizer'' baseline achieves 99\% accuracy, matching QACGAN, demonstrating that the energy term contributes nothing beyond the corrected training protocol. The present results therefore do not constitute evidence for a quantum advantage, nor for a uniquely quantum inductive bias in the studied setting.

\subsection{Why Did the VQE Approach Fail to Show Advantage?}

Several factors explain why VQE regularization provides no benefit in this setting:

\begin{enumerate}
    \item \textbf{Trivial Hamiltonian design.} The linear parameterization ($h_{c,i} = 0.1 + 0.01c$) encodes no meaningful structure. The ``quantum'' component is essentially computing a class-dependent scalar through a $\sim$200$\times$ more expensive pathway that an MLP or fixed bias can trivially replicate.
    
    \item \textbf{Insufficient problem complexity.} MNIST classification is too simple to require sophisticated regularization. With proper training protocols, even unregularized ACGAN achieves near-perfect class consistency.
    
    \item \textbf{Gradient interference.} The VQE energy pathway introduces additional variance (QACGAN FID std = $\pm 8.0$ vs.\ classical variants $\pm 1$--$4$), suggesting the quantum gradient signal may actually destabilize training.
\end{enumerate}

\subsection{Critical Limitations}

\textbf{Simulator-only evaluation.} All results were obtained on a noiseless statevector simulator. Real quantum hardware introduces noise, decoherence, and finite sampling that could alter behavior; though given the negative result, there is no strong motivation to pursue hardware evaluation.

\textbf{Limited problem domain.} MNIST may be too simple to reveal any potential benefit of quantum-derived regularization. However, extending to more complex domains would require justified hypotheses about \emph{why} quantum formulations might help, which the current negative result does not support.

\textbf{Computational overhead.} The $\sim$200$\times$ slowdown for VQE computation (Python-level simulator overhead) represents a severe practical limitation even if performance benefits existed.

\subsection{Implications for Quantum Machine Learning Research}

This negative result carries important methodological lessons:

\begin{enumerate}
    \item \textbf{Ablation studies are essential.} Claims of ``quantum-enhanced'' performance require controlled comparisons against classical baselines with matched capacity and structure. Without such comparisons, apparent quantum effects may simply reflect the presence of any auxiliary signal.
    
    \item \textbf{Trivial Hamiltonians provide no benefit.} If the quantum component encodes no meaningful problem structure, it cannot be expected to outperform classical surrogates that produce equivalent outputs.
    
    \item \textbf{Implementation details matter.} The difference between 87.8\% and 99\% accuracy was attributable to label smoothing and gradient accumulation, not VQE regularization. Careful control of training protocols is necessary for valid comparisons.
\end{enumerate}

\subsection{Future Directions}

Given the negative result, future work should focus on:

\begin{itemize}
    \item \textbf{Non-trivial Hamiltonians:} Designing problem-aware Hamiltonians that encode meaningful structure (e.g., data-dependent couplings, learned parameterizations) that classical surrogates cannot trivially replicate.
    
    \item \textbf{Complex domains:} Testing on problems where auxiliary regularization demonstrably helps classical GANs, providing a clearer baseline against which quantum alternatives can be compared.
    
    \item \textbf{Theoretical justification:} Developing principled arguments for \emph{why} quantum-derived regularization might provide advantages in specific settings before conducting empirical evaluations.
\end{itemize}

\section{Conclusion}
We presented an exploratory investigation of VQE-inspired energy terms as auxiliary regularization signals in GANs, accompanied by a rigorous ablation study against classical baselines.

\textbf{Main finding (negative result):} The VQE component provides \emph{no measurable causal benefit} beyond what is achievable with trivial classical alternatives. All tested variants, MLP-based energy, learned per-class bias, random noise, and pure unregularized ACGAN, achieved equivalent performance ($\approx$99\% accuracy, FID 18--21) when trained under fair conditions.

The initial observation that QACGAN achieved 99--100\% accuracy versus 87.8\% for ACGAN was attributable to implementation differences (label smoothing, gradient accumulation) rather than the VQE term itself. When these factors are controlled, the quantum component contributes nothing.

\textbf{Methodological contribution:} This work demonstrates the importance of ablation-driven methodology in quantum machine learning research. It shows that apparent ``quantum effects'' can arise from confounded comparisons and that rigorous classical baselines are essential before claiming any quantum-derived benefit.

\textbf{Negative results matter:} We publish this negative result because it provides actionable guidance for the QML community: (1)~ablation studies are non-negotiable, (2)~trivial Hamiltonians cannot be expected to outperform classical alternatives, and (3)~implementation details can dominate apparent effects. The primary contribution of this work lies not in performance gains, but in clarifying which hybrid quantum--classical mechanisms do not yield causal benefits under controlled evaluation. By demonstrating that a seemingly plausible hybrid quantum--classical mechanism does not survive controlled ablation, this work highlights both the necessity and the value of negative results in the evaluation of hybrid generative models.

\section*{Acknowledgments}
The author would like to express sincere gratitude to Prof.\ Roman Šenkeřík 
for his supervision and support during the preparation of this work.

\bibliographystyle{plainnat}
\bibliography{references}

\end{document}